\pdfoutput=1

\documentclass[11pt]{article}

\usepackage[final]{acl}

\usepackage{times}
\usepackage{latexsym}
\usepackage{enumitem}
\usepackage[T1]{fontenc}

\usepackage[utf8]{inputenc}

\usepackage{microtype}

\usepackage{inconsolata}

\usepackage{graphicx}
\usepackage{booktabs}
\usepackage{amssymb}
\title{Image captioning in different languages}

\author{Emiel van Miltenburg\\
 Tilburg University\\
  \texttt{C.W.J.vanMiltenburg@tilburguniversity.edu}}

\begin{document}
\maketitle
\begin{abstract}
This short position paper provides a manually curated list of non-English image captioning datasets (as of May 2024). Through this list, we can observe the dearth of datasets in different languages: only 23 different languages are represented. With the addition of the Crossmodal-3600 dataset \cite[36 languages]{thapliyal-etal-2022-crossmodal} this number increases somewhat, but still this number is small compared to the $\pm$~500 institutional languages that are out there. This paper closes with some open questions for the field of Vision \& Language.
\end{abstract}

\section{Introduction}
Image captioning datasets have been developed for several different languages, but only a handful (mostly English datasets) have received attention in the NLP community. This research note aims to provide an overview of the different image captioning datasets that have been described in the literature. This list is the result of monitoring Google Scholar alerts for the last eight years using the keywords "MS COCO" and "Flickr30K" (the prototypical English captioning datasets), and manually curating a list of papers that describe image captioning work in languages other than English.

\subsection{Earlier work}
An earlier version of this list appeared in the PhD thesis of \citet{van2019pragmatic}. The list shows the language of the dataset, the source of the images (either an existing dataset such as Flickr8/30K and MS COCO or the name of the newly introduced dataset, whether the captions were translated from English or independently collected, and a reference to the paper where the dataset was introduced.

Next to individual references, I also relied on the survey by \citet{jayaswal_comprehensive_2023} to get an overview of image captioning work for Indian languages (i.e.\ \textit{Assamese, Bengali/Bangla, Hindi, Punjabi, Tamil, Telugu}, and \textit{Urdu}). Although the survey itself does not contain a list of existing datasets, the references provided a useful starting point to add all Indian captioning datasets to the list.

\subsection{Additions}
Given the creation history of this overview, I probably have overlooked some datasets. Let me know if you have developed a captioning dataset in a language other than English.

\begin{table*}
	\small
	\centering
	\begin{tabular}{lllll}
		\toprule
		Language & Source & T & I & Citation\\
		\midrule
        Arabic & JEEM & & $\checkmark$ & \citealt{kadaoui2025jeemvisionlanguageunderstandingarabic}\\
        Arabic & ArabicFashionData & & $\checkmark$ & \citealt{s23083783}\\
        Arabic & MS COCO* & $\checkmark$& &\citealt{eljundi2020resources}\\
        Arabic & MS COCO & $\checkmark$ &  & \citealt{2023arXiv231108844M}\\
        Arabic & MS COCO* &  & $\checkmark$ & \citealt{2023arXiv231108844M}\\
        Arabic & MS COCO* & $\checkmark$ &  & \citealt{Al-muzaini2018}\\
        Arabic & Flickr8K* & $\checkmark$ &  & \citealt{Al-muzaini2018}\\
		Assamese & Assamese-news & & $\checkmark$ & \citealt{das-singh-2021-image,das2022assamese}\\
        Assamese & Flickr30K & $\checkmark$ & & \citealt{nath-etal-2022-image}\\
        Assamese & $\hookrightarrow$ revised & $\checkmark$ & & \citealt{choudhury-etal-2023-image}\\
        Assamese & MS COCO & $\checkmark$ & & \citealt{nath-etal-2022-image}\\
        Assamese & $\hookrightarrow$ revised & $\checkmark$ & & \citealt{choudhury-etal-2023-image}\\
        Bengali & BanglaLekha & & $\checkmark$ & \citealt{RAHMAN2019636}\\
        Bengali & Bornon & & $\checkmark$ & \citealt{shah2021bornon}\\
        Bengali & Flickr8K & $\checkmark$ & & \citealt{shah2021bornon,Humaira2021}\\
        Bengali & Flickr8K & $\checkmark$ & & \citealt{khan-etal-2022-ban}\\
        Bengali & BNLIT & & $\checkmark$ & \citealt{IJAIN499}\\
        Bengali & Flickr30K & $\checkmark$ & & \citealt{10099345}\\
        Bengali & Flickr30K & $\checkmark$ & & \citealt{https://doi.org/10.17632/rrv8pbxrxv.1}\\
        Brazilian Portuguese & Flickr30K & $\checkmark$ & $\checkmark$ & \citealt{viridiano-etal-2024-framed-multi30k}\\
        Chinese & Flickr8K & $\checkmark$ & $\checkmark$ & \citealt{li2016adding} \\
        Chinese & Flickr30K & $\checkmark$ & $\checkmark$ & \citealt{DBLP:journals/corr/abs-1708-04390}\\
		Chinese & MS COCO* & $\checkmark$ & $\checkmark$ & \citealt{DBLP:journals/corr/abs-1805-08661}\\
        Chinese & AI Challenger/ICC & & $\checkmark$ & \citealt{DBLP:journals/corr/abs-1711-06475}\\
        Chinese & Thangka & & $\checkmark$ & \citealt{jimaging9080162}\\
		Czech & Flickr30K & $\checkmark$ & & \citealt{barrault-EtAl:2018:WMT}\\
		Dutch & Flickr30K* & & $\checkmark$ & \citealt{vanmiltenburg-elliott-vossen:2017:INLG2017}\\
		Dutch & MS COCO* & & $\checkmark$ & \citealt{miltenburg2018DIDEC}\\
		French & MS COCO* & $\checkmark$ & & \citealt{rajendran2016bridge}\\
		French & Flickr30K & $\checkmark$ & & \citealt{elliott-EtAl:2017:WMT}\\
		German & MS COCO* & $\checkmark$ & & \citealt{rajendran2016bridge}\\
		German & MS COCO* & $\checkmark$ & & \citealt{hitschlerETAL:16} \\
		German & Flickr30K & $\checkmark$ & $\checkmark$ & \citealt{elliott-EtAl:2016:VL16}\\
		German & IAPR-TC12 & & $\checkmark$ & \citealt{grubinger2006iapr}\\
        Hindi & Flickr8K & $\checkmark$ & & \citealt{9223087}\\
        Hindi & MS COCO & $\checkmark$ & & \citealt{dhir2019deep,10.1145/3432246,MISHRA2021107114}\\	
        Hindi & Visual Genome & $\checkmark$ & & \citealt{hindi-visual-genome:2019}\\
        Indonesian & Flickr8K & $\checkmark$ & & \citealt{10497459}\\
        Indonesian & MS COCO & $\checkmark$ & & \citealt{10.1063/5.0118155}\\
        Japanese & UIUC Pascal & $\checkmark$ & & \citealt{funaki2015image}\\ 
		Japanese & MS COCO* & & $\checkmark$ & \citealt{miyazaki2016cross}\\
		Japanese & MS COCO & & $\checkmark$ & \citealt{yoshikawa2017stair}\\
        Persian & ParsVQA-Caps & & $\checkmark$ & \citealt{mobasher_parsvqa-caps_2022}\\
        Portuguese & \#{}PraCegoVer & & $\checkmark$ & \citealt{santos_pracegover_2021}\\
        Punjabi & Punjabi Tribune & & $\checkmark$ & \citealt{Kaur_Josan_Kaur_2021}\\
		Spanish & IAPR-TC12 & $\checkmark$ & & \citealt{grubinger2006iapr}\\
		\bottomrule
	\end{tabular}
	\caption{Image description datasets available in languages other than English (A-S), with an indication of their source, and whether the descriptions were \textbf{T}ranslated or \textbf{I}ndependently collected. Asterisks indicate that the data is a subset of the original dataset.
	Flickr8K is the predecessor of Flickr30K, see \protect\citealt{hodosh2013framing}. For some recources multiple references are given since it is not clear which is the `main' reference. In the case of Assamese, \citet{choudhury-etal-2023-image} manually corrected translation errors from the datasets provided by \citet{nath-etal-2022-image}, resulting in a revised (but highly similar) dataset.}
	\label{table:otherlanguages}
\end{table*}

\begin{table*}
	\small
	\centering
	\begin{tabular}{lllll}
		\toprule
		Language & Source & T & I & Citation\\
		\midrule
        Tamil & MS COCO & $\checkmark$ & & \citealt{9848810}\\
        Tamil & MS COCO & $\checkmark$ & & \citealt{b_multilingual_2023}\\
        Tamil & Flickr8K & $\checkmark$ & & \citealt{b_multilingual_2023}\\
        Tamil & Flickr30K & $\checkmark$ & & \citealt{b_multilingual_2023}\\
        Telugu & MS COCO & $\checkmark$ & & \citealt{b_multilingual_2023}\\
        Telugu & Flickr8K & $\checkmark$ & & \citealt{b_multilingual_2023}\\
        Telugu & Flickr30K & $\checkmark$ & & \citealt{b_multilingual_2023}\\
        Telugu & Flickr8K & $\checkmark$ & & \citealt{9441908}\\
        Thai & HouseDefects & & $\checkmark$ & \citealt{Jaruschaimongkol2023automatic}\\
        Thai & Thai-Food & & $\checkmark$ & \citealt{9960246}\\
        Thai & Thai-Travel & & $\checkmark$ & \citealt{9960246}\\
        Thai & Flickr30K & $\checkmark$ & & \citealt{9960246}\\
		Turkish & Flickr8K & & $\checkmark$ & \citealt{unal2016tasviret}\\
        Turkish & MS COCO & $\checkmark$ & & \citealt{9988025}\\
        Turkish & MS COCO & $\checkmark$ & & \citealt{10286693}\\
        Turkish & Tiny TR-cap & & $\checkmark$ & \citealt{MEMIS2025102009}\\
        Ukrainian & Flickr30K & $\checkmark$ & & \citealt{saichyshyna-etal-2023-extension}\\
        Urdu & Flickr8K* & $\checkmark$ & & \citealt{ilahi2020efficient}\\
        Urdu & Flickr8K* & $\checkmark$ & & \citealt{afzal2023generative}\\
        Vietnamese & MS COCO* & & $\checkmark$ & \citealt{DBLP:journals/corr/abs-2002-00175}\\
        Vietnamese & KTVIC & & $\checkmark$ & \citealt{pham_ktvic_2024}\\
        Vietnamese & UIT-OpenViIC & & $\checkmark$ & \citealt{2023arXiv230504166B}\\
		\bottomrule
	\end{tabular}
	\caption{Image description datasets available in languages other than English (T-Z), with an indication of their source, and whether the descriptions were \textbf{T}ranslated or \textbf{I}ndependently collected. Asterisks indicate that the data is a subset of the original dataset.
	Flickr8K is the predecessor of Flickr30K, see \protect\citealt{hodosh2013framing}. For some recources multiple references are given since it is not clear which is the `main' reference.}
	\label{table:otherlanguages2}
\end{table*}

\section{The list}
Tables~\ref{table:otherlanguages} and \ref{table:otherlanguages2} provide a list of non-English captioning datasets.\footnote{The Crossmodal-3600 dataset \cite[36 languages]{thapliyal-etal-2022-crossmodal} is not listed in the table because it would result in too much repetition.}$^{,}$\footnote{Some authors opt not to translate the dataset but instead use an English-language captioning model and then translate the results (e.g. \citealt{10.1007/978-3-030-63119-2_64}). These works are also not in the table.} Some observations:
\begin{itemize}
    \item Starting with the most obvious observation: although I made an effort to be inclusive and bookmark all non-English captioning datasets I came across, the table is still not very diverse. Only a handful of languages have researchers actively producing image captioning datasets.

    \item Many datasets have been created through (automatic) translation of English captions, which is relatively cheap compared to independently collected captions. The downside, of course, is that these datasets still capture a Western perspective, even if the dataset has been written in a non-Western language. 

    \item Although Tables~\ref{table:otherlanguages} and \ref{table:otherlanguages2} differentiate between translated and independently collected descriptions, in practice this distinction may be blurred by annotators providing manual corrections of automatic translations, sometimes replacing the translations altogether.

    \item Some datasets are very specific to a particular use case, for example the Arabic Fashion dataset \cite{s23083783}, the Chinese Thangka dataset \cite{jimaging9080162}, and the Thai house defects dataset \cite{Jaruschaimongkol2023automatic}. These may not be useful for general image captioning tasks, but may still help to ground multi-modal language models.

    \item Most of the papers in Table~\ref{table:otherlanguages} have not been published at ACL workshops (or their Computer Vision equivalents), meaning that these are easily missed if you aren't explicitly looking for them.

    \item There are six languages in this list that are not in the Crossmodal-3600 dataset. These are: Assamese, Brazilian Portuguese, Portuguese, Punjabi, Tamil, and Urdu. This brings the total number of languages with available caption data to 36 + 6 = 42.
\end{itemize}

\section{Research questions}
Browsing the list above, many research questions come to mind. For example:

\begin{description}[wide]
    \item[Assessment] How well do existing Vision \& Language models perform on different datasets, such as Assamese-news, BNLIT, or Thai-Food/Travel? And, since any caption task inherently makes (implicit) assumptions about the intended audience, from which perspective should we assess these models?

    \item[Benefiting from multi-lingual data] How can we benefit from multi-lingual datasets? And how do we overcome the challenges brought by the imbalance in available training data?
    
    \item[Inclusion and Community Building] How inclusive are Vision \& Language research and citation practices? Are we doing enough to accommodate and support speakers of underrepresented languages? 
\end{description}

The last point clearly stands out from the previous two, in that it does not only require desk research but also carrying out community work, talking to different stakeholders, and bridging the gap between different cultures. This seems most important to make Vision \& Language models work for everyone, yet this work is not strongly incentivized.

\section{Discussion}
It is often said that there are about 7,000 languages in the world, which may lead us to think that we should also develop NLP technology for all of those languages. But \citet{bird-2024-must} reminds us that beyond the 500 institutional languages, there are 6,500 oral languages that may not require technological solutions such as image captioning.\footnote{See \citet{bird-yibarbuk-2024-centering} for a more in-depth discussion of these numbers.} Instead, the \emph{contact languages} used between `outsiders' and speakers of those oral languages (that overlap with the 500 institutional languages) should be the target language for NLP technology. 

\section{Conclusion}
This note provides an overview of non-English captioning datasets. This is not only useful for those who would like to train a captioning model for the different languages that are listed in the table, but it hopefully also increases the visibility of the authors who have been working to promote image captioning technology for their own languages. It is also clear that much work remains to be done, not only developing datasets and models for different languages and cultures, but also in terms of building a community that embraces cultural diversity.

\bibliography{custom}
\end{document}